\documentclass[10pt, a4paper]{article}
\usepackage{lrec}
\usepackage[dvipdfmx]{graphicx}
\usepackage{tabularx}
\usepackage{soul}
\usepackage{hyperref}
\usepackage{xstring}
\usepackage{times}
\usepackage{latexsym}
\usepackage{breakurl}
\usepackage{amsfonts}
\usepackage{amsmath}
\usepackage{mathtools}
\usepackage{bm}
\usepackage{lscape}
\usepackage{algorithm,algpseudocode}
\usepackage{xcolor}
\usepackage{ulem}
\usepackage{caption}
\usepackage{setspace}
\usepackage{booktabs}

\usepackage{CJKutf8}
\usepackage[utf8]{inputenc}
\usepackage[T1]{fontenc}
\newcommand{\ja}[1]{\begin{CJK}{UTF8}{ipxm}#1\end{CJK}}

\newcommand{\secref}[1]{\StrGobbleRight{\getrefnumber{#1}}{1}}

\title{JParaCrawl: A Large Scale Web-Based English-Japanese Parallel Corpus}

\name{Makoto Morishita, Jun Suzuki, and Masaaki Nagata}
\address{NTT Communication Science Laboratories, NTT Corporation\\
  2-4 Hikaridai, Seika-cho, Soraku-gun, Kyoto, 619-0237, Japan\\
  {\tt \{makoto.morishita.gr, masaaki.nagata.et\}@hco.ntt.co.jp}\\
  {\tt jun.suzuki@ecei.tohoku.ac.jp}}

\abstract{
Recent machine translation algorithms mainly rely on parallel corpora.
However, since the availability of parallel corpora remains limited, only some resource-rich language pairs can benefit from them.
We constructed a parallel corpus for English-Japanese, for which the amount of publicly available parallel corpora is still limited.
We constructed the parallel corpus by broadly crawling the web and automatically aligning parallel sentences.
Our collected corpus, called JParaCrawl, amassed over 8.7 million sentence pairs.
We show how it includes a broader range of domains and how a neural machine translation model trained with it works as a good pre-trained model for fine-tuning specific domains.
The pre-training and fine-tuning approaches achieved or surpassed performance comparable to model training from the initial state and reduced the training time.
Additionally, we trained the model with an in-domain dataset and JParaCrawl to show how we achieved the best performance with them.
JParaCrawl and the pre-trained models are freely available online for research purposes.
\\ \newline
}

\begin{document}

\maketitleabstract

\section{Introduction}
Since current machine translation (MT) approaches are mainly data-driven, one key bottleneck has been the lack of parallel corpora.
This problem continues with the recent neural machine translation (NMT) architecture.
As the amount of training data increases, NMT performance improves \cite{sennrich19acl}.

Our goal is to create large parallel corpora to/from Japanese.
In our first attempt, we focused on the English-Japanese language pair.
Currently, ASPEC is the largest publicly available English-Japanese parallel corpus \cite{nakazawa16aspec}, which contains 3.0 million sentences for training.
Unfortunately, this is relatively small compared to such resource-rich language pairs as French-English\footnote{Currently the largest French-English parallel corpus is ParaCrawl v5, which contains 51.3 million training data.}.
Also, available domains remain limited.
We address this problem, which hinders the progress of English-Japanese translation research,
by crawling the web to mine for English-Japanese parallel sentences.

Current NMT training requires a great deal of computational time, which complicates running experiments with few computational resources.
We alleviate this problem by providing NMT models trained with our corpus.
Since our web-based parallel corpus contains a broad range of domains, it might be used as a pre-trained model and fine-tuned with a domain-specific parallel corpus.

The following are the contributions of this paper:
\begin{itemize}
    \item We released the largest freely available web-based English-Japanese parallel corpus: JParaCrawl.
    \item We also released NMT models trained with JParaCrawl for further fine-tuning.
\end{itemize}

The rest of the paper is organized as follows.
In Section~\secref{sec:related_work}, we introduce previous work that tried to create parallel corpora by mining parallel texts.
In Section~\secref{sec:jparacrawl}, we show how we created JParaCrawl.
We conducted experiments to show how effectively our corpus and pre-trained models worked with typical NMT training settings.
These experimental settings and results are shown in Section~\secref{sec:experiments}.
Finally, we conclude with a brief discussion of future work in Section~\secref{sec:conclusion}.

JParaCrawl and the NMT models pre-trained with it are freely available online\footnote{\url{http://www.kecl.ntt.co.jp/icl/lirg/jparacrawl/}} for research purposes\footnote{This research is based on JParaCrawl v1.0. During the reviewing process, we released v2.0, which contains more than 10 million sentence pairs.}.

\section{Related Work}
\label{sec:related_work}
One typical kind of source for parallel texts is the documents of international organizations.
Europarl \cite{koehn05europarl} is an example of the early success of creating a large parallel corpus by automatically aligning parallel texts from the proceedings of the European Parliament.
The United Nations Parallel Corpus \cite{ziemski16un} is another similar example that was created from UN documents.
These texts were translated by professionals, and aligning the documents is easy because they often have such meta-information as the identities of speakers, although their domains and language pairs are limited.

Another important source of parallel texts is the web.
\newcite{uszkoreit10mining} proposed a large-scale distributed system to mine parallel text from the web and books.
\newcite{smith13dirtcheap} proposed an algorithm that creates a parallel corpus by mining Common Crawl\footnote{\url{https://commoncrawl.org/}}, which is a free web crawl archive.
\newcite{schwenk19wikimatrix} mined Wikipedia and created a parallel corpus of 1,620 language pairs.
The web, which includes a broad range of domains and many language pairs, is rapidly and continually growing.
Thus, it has huge potential as a source of parallel corpora, although identifying correct parallel sentences is difficult.

Our work was inspired by the recent success of the ParaCrawl\footnote{\url{https://paracrawl.eu/}} project, which is building parallel corpora by crawling the web.
Their objective is to build parallel corpora to/from English for the 24 official languages of the European Union.
They released an earlier version of the corpora, and this version is already quite large\footnote{For example, the English-German sentence pairs in the v5.0 release already exceed 36 million.}.
This early release has already been used for previous WMT shared translation tasks for some language pairs \cite{bojar18wmt,barrault19wmt}, and task participants reported that ParaCrawl significantly improved the translation performance when it was used with a careful corpus cleaning technique \cite{microsoft18wmt}.
We extend this work to mine English-Japanese parallel sentences.

\section{JParaCrawl}
\label{sec:jparacrawl}

\begin{figure*}[t]
    \centering
    \includegraphics[scale=0.85, clip]{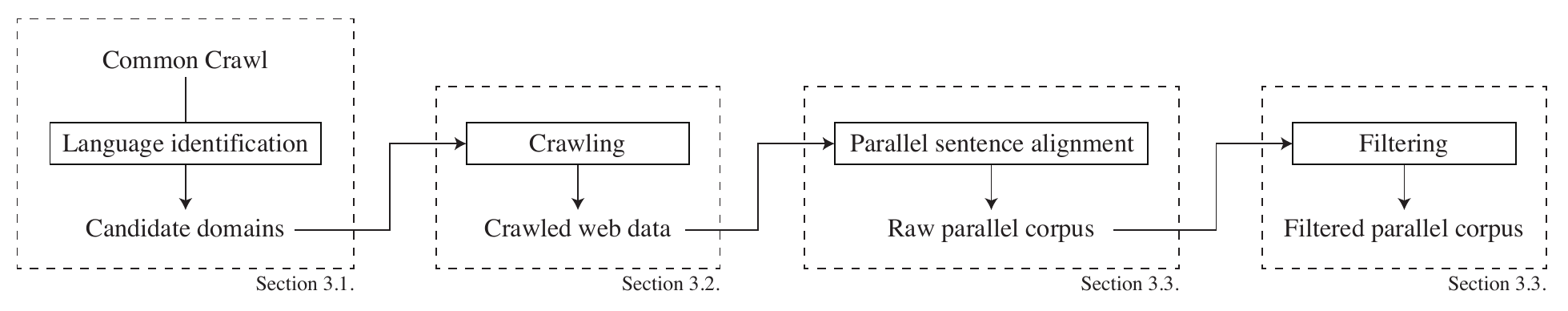}
    \caption{Flowchart showing how we mined parallel sentences from web}
    \label{fig:corpus_flowchart}
\end{figure*}

To collect English-Japanese parallel sentences from the web, we took a similar approach to that used in the ParaCrawl project.
Figure~\ref{fig:corpus_flowchart} shows how we mined parallel corpora from the web.
First, we selected candidate domains to crawl that might contain English-Japanese parallel sentences from Common Crawl data (Section~\secref{sec:crawling_domain_selection}).
Then, we crawled the candidate domains (Section~\secref{sec:crawling}).
Finally, we aligned the parallel sentences from the crawled data and filtered out noisy sentence pairs (Section~\secref{sec:bitext_alignment}).

\subsection{Crawling Domain Selection}
\label{sec:crawling_domain_selection}
One key to creating a parallel corpus from the web is to decide which websites to crawl (i.e., finding candidate domains that contain a large number of parallel sentences).
To select candidate domains, we first identified the languages of all the Common Crawl text data by {\tt CLD2}\footnote{\url{https://github.com/CLD2Owners/cld2}} and counted how much Japanese or English data each domain has.
We used an {\tt extractor}\footnote{\url{https://github.com/paracrawl/extractor}} provided by the ParaCrawl project for creating language statistics for each domain.

If the amount of each language is balanced, the website might contain good parallel sentences.
Thus, we used the ratio between Japanese and English as a criterion to select the candidate domains.
We ranked all websites based on their language ratios and listed the top 100,000 domains to crawl.

\subsection{Crawling the Web}
\label{sec:crawling}
The next step was to crawl the candidate websites to mine for parallel sentences.
Since the crawled data stored on Common Crawl may not contain entire websites or might be outdated, we recrawled the websites ourselves
with {\tt HTTrack}\footnote{\url{http://www.httrack.com/}}.
We stopped crawling a website if we could not crawl it in 24 hours.
In this experiment, we focused on text data; future work will include other formats, such as PDFs.
After crawling 100,000 domains, our crawled data exceeded 8.0 TB with gzip compression.

\subsection{Bitext Alignment}
\label{sec:bitext_alignment}

\begin{table}[tb]
\centering
\begin{tabular}{lrr}
\toprule
     & \textbf{\# sentences} & \textbf{\# words}\\ \midrule
Raw & 27,724,570 & 594,013,165\\
Filtered & 8,763,995 & 196,084,272\\
\bottomrule
\end{tabular}
\caption{Number of collected sentences and words on English side in JParaCrawl corpus}
\label{tab:corpus_stats}
\end{table}

Next, we describe how we mined parallel text from the crawled web data.

We crawled a large number of websites, but some were too small to mine for parallel sentences.
Therefore, we filtered out those domains whose compressed archive size was less than 1 MB, and only 39,936 domains remained.

To align parallel sentences, we used the {\tt Bitextor} toolkit\footnote{\url{https://github.com/bitextor/bitextor}} provided by the ParaCrawl project.
Our experiment was based on version 7.0, and we fixed several components for Japanese sentences.
To extract text from HTML files, we used {\tt extractontent}\footnote{\url{https://github.com/yono/python-extractcontent}}, which was developed for Japanese text.
We used {\tt split-sentences.perl}\footnote{\url{https://github.com/moses-smt/mosesdecoder/blob/master/scripts/ems/support/split-sentences.perl}} contained in the Moses toolkit \cite{koehn07moses} to split a text into sentences.
We fixed the script to deal with Japanese end-of-sentence tokens.

There are two primary approaches to align parallel text.
One algorithm uses a bilingual lexicon to generate crude translations along with such other features as sentence length to find the best sentence pair \cite{varga05hunalign}.
The other algorithm uses an external MT system to translate one language into the other and find a sentence pair that maximizes BLEU scores \cite{sennrich11bleualign,papineni02bleu}.
Since the latter approach needs more computational resources for external MT, we used the bilingual lexicon-based algorithm.
We used the EDR English-Japanese dictionary as a bilingual lexicon \cite{miyoshi96edr}.

Table~\ref{tab:corpus_stats} shows the number of collected parallel sentences and words.
After the bitext alignment process, we mined over 27 million parallel sentences.
However, these collected sentences contained many noisy pairs.
Therefore, we filtered the corpus with {\tt Bicleaner}\footnote{\url{https://github.com/bitextor/bicleaner}} \cite{prompsit18bicleaner}.
The filtering model was trained with our in-house English-Japanese parallel corpora.
After removing sentence pairs whose scores were lower than 0.5, we retained around 8.7 million sentences.

As our initial corpus release, we open-filtered 8.7 million parallel sentences to the public.
However, we still found noisy sentence pairs that should have been filtered out.
Future work will improve our filtering algorithm.

\section{Experiments}
\label{sec:experiments}

\begin{figure*}[t]
    \centering
    \includegraphics[scale=0.85, clip]{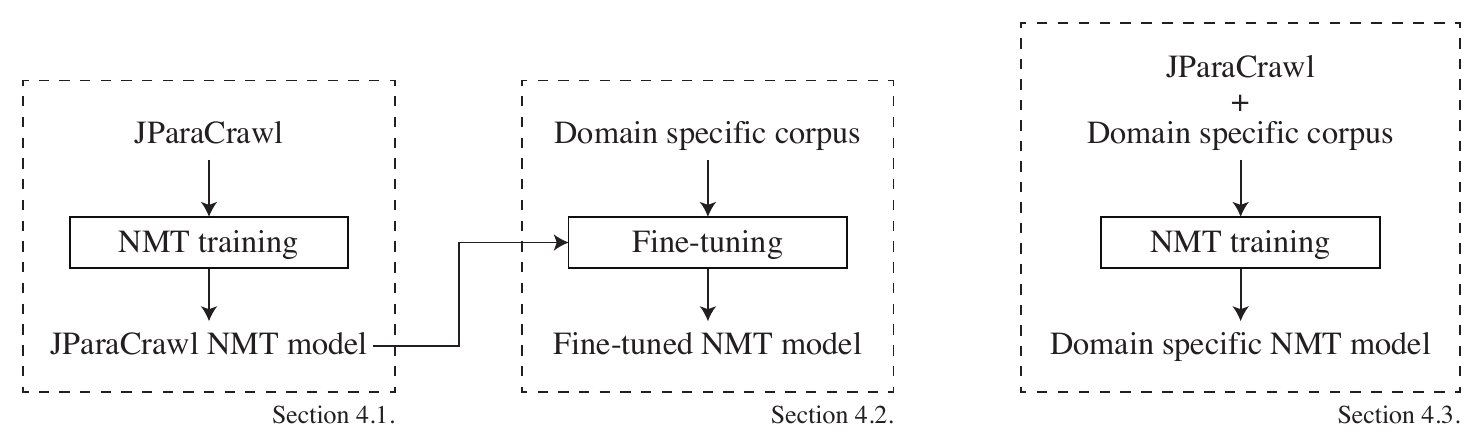}
    \caption{Overview of experiments: We first trained an NMT model with JParaCrawl in Section~\protect\secref{sec:train_with_jparacrawl}\ and fine-tuned it with a domain-specific corpus in Section~\protect\secref{sec:pretrain_models}. We also trained a model for a specific domain without fine-tuning in Section~\protect\secref{sec:train_specific_domain}.}
    \label{fig:training_flowchart}
\end{figure*}

We carried out three experiments to show how our corpus effectively works with typical NMT training settings.
Figure~\ref{fig:training_flowchart} shows an overview of the experiments.
In Section~\secref{sec:train_with_jparacrawl}, we trained an NMT model with JParaCrawl.
Then, we pre-trained an NMT model with JParaCrawl and fine-tuned it with an existing corpus in Section~\secref{sec:pretrain_models}.
Finally, we trained a model for a specific domain with JParaCrawl and other corpora without fine-tuning in Section~\secref{sec:train_specific_domain}.
In the following experiments, we used a filtered JParaCrawl corpus that contained 8.7 million parallel sentences.

\subsection{Training NMT with JParaCrawl}
\label{sec:train_with_jparacrawl}
In this section, we trained NMT models with JParaCrawl and tested the models on several test sets to see how our corpus covers a broader range of domains.

\subsubsection{Experimental Settings}
\label{sec:train_with_jparacrawl_settings}
\begin{table}[t]
\centering
\begin{tabular}{lrr}
\toprule
\textbf{Data} & \textbf{\# sentences} & \textbf{\# words}\\ \midrule
ASPEC & 1,812 & 39,573\\
JESC & 2,000 & 13,617\\
KFTT & 1,160 & 22,063\\
IWSLT  & 1,194 & 20,367\\
\bottomrule
\end{tabular}
\caption{Number of sentences and words on English side in test sets}
\label{tab:test_set_corpus_stats}
\end{table}

\begin{table}[t]
\centering
\begin{tabular}{lrr}
\toprule
\textbf{Data} & \textbf{\# sentences} & \textbf{\# words}\\ \midrule
ASPEC & 3,008,500 & 68,929,413\\
JESC & 2,797,388 & 19,339,040\\
KFTT & 440,288 & 9,737,715 \\
IWSLT & 223,108 & 3,877,868\\
\bottomrule
\end{tabular}
\caption{Number of sentences and words on English side in training sets. The original version of ASPEC contains 3.0 million sentences, but we used only the first 2.0 million sentences for training based on previous work \protect\cite{neubig14wat}.}
\label{tab:train_set_corpus_stats}
\end{table}

\paragraph{Data}
To see how our corpus covers a broader range of domains, we used four test sets: scientific paper excerpts (ASPEC, \newcite{nakazawa16aspec}), movie subtitles (JESC, \newcite{pryzant17jesc}), texts on Wikipedia articles related to Kyoto (KFTT, \newcite{neubig11kftt}), and TED talks (tst2015, provided for the IWSLT 2017 evaluation campaign, \newcite{cettolo17iwslt}, \newcite{cettolo12wit3}).
Table~\ref{tab:test_set_corpus_stats} shows the details of the test sets.
During training, we used the ASPEC dev set as a validation set.

We preprocessed the data with {\tt sentencepiece} \cite{kudo18sentencepiece} to split the sentences into subwords.
We set the vocabulary size to 32,000 and removed sentences whose length exceeded 250 subwords from the training data.
Since JParaCrawl was NFKC-normalized, we also normalized the test sets.

For comparison, we trained NMT models with domain-specific bitexts.
Table~\ref{tab:train_set_corpus_stats} shows the number of sentences and words in the domain-specific training sets.
Since the sentences in ASPEC are ordered by their alignment confidence scores, the former sentences tended to be clean and the latter might contain noisy sentence pairs.
Based on previous work \cite{neubig14wat}, we used only the first 2.0 million sentences for training, although the original ASPEC corpus contained 3.0 million sentences.

\paragraph{NMT Models}
We trained an NMT model with {\tt fairseq} \cite{ott19fairseq}.
Our model was based on Transformer \cite{vaswani17transformer}.
We trained three models for each direction by varying the hyper-parameters: small, base, and big settings.
The base and big settings are based on \newcite{vaswani17transformer}.
For the base settings, we used an encoder/decoder with six layers.
We set their embedding size to 512 and their feed-forward embedding size to 2048.
We used eight attention heads for both the encoder and the decoder.
For the big settings, we changed their embedding size to 1024 and their feed-forward embedding size to 4096.
We also used 16 attention heads for both the encoder and the decoder.
For the small settings, we used the same settings as for the base settings, except we changed the number of heads to four.
For all the settings, we used dropout with a probability of 0.3 \cite{srivastava14dropout}.
We used big settings for ASPEC and JESC, base settings for KFTT, and small settings for IWSLT.

As an optimizer, we used Adam with $\alpha=0.001$, $\beta_{1}=0.9$, and $\beta_{2}=0.98$.
We used a root-square decay learning rate schedule with a linear warmup of 4000 steps \cite{vaswani17transformer}.
We clipped gradients to avoid exceeding their norm 1.0 to stabilize the training \cite{pascanu13clipping}.
For the base and small settings, each mini-batch contained about 5,000 tokens (subwords), and we accumulated the gradients of 64 mini-batches for updates \cite{ott18scaling}.
For the big settings, we set the mini-batch size to 2,000 tokens and accumulated 160 mini-batches for updates.
We trained the model with 24,000 iterations, saved the model parameters every 200 iterations, and averaged the last eight models.
To achieve maximum performance with the latest GPUs, we used mixed-precision training \cite{micikevicius18mixed}.
When decoding, we used a beam search with a size of six and length normalization by dividing the scores by their lengths.

We slightly changed the settings for the in-domain baseline training because some of the above settings are inappropriate for smaller datasets.
For IWSLT, we accumulated 16 mini-batches per update instead of 64.
Since we confirmed that the model had already converged based on the validation loss, we stopped training at 20,000 iterations for all the in-domain baselines.

\paragraph{Evaluation}
To evaluate the performance, we calculated the BLEU scores \cite{papineni02bleu} with {\tt sacreBLEU} \cite{post18sacrebleu}.
Since {\tt sacreBLEU} does not internally tokenize Japanese text, we tokenized both the hypothesis and reference texts using MeCab\footnote{\url{https://taku910.github.io/mecab/}} with an IPA dictionary when evaluating the English-Japanese translations.

\subsubsection{Experimental Results and Analysis}

\begin{table*}[t]
\centering
\begin{tabular}{lrrrrrrrrrr}
\toprule
\multicolumn{1}{l}{\bf Test data} & \multicolumn{5}{c}{\bf English-to-Japanese} & \multicolumn{5}{c}{\bf Japanese-to-English}\\
\cmidrule(lr){2-6}
\cmidrule(lr){7-11}&
\multicolumn{1}{r}{\bf In-domain} & \multicolumn{2}{r}{\bf JParaCrawl} & \multicolumn{2}{r}{\bf Fine-tuning} &
\multicolumn{1}{r}{\bf In-domain} & \multicolumn{2}{r}{\bf JParaCrawl} & \multicolumn{2}{r}{\bf Fine-tuning} \\ \midrule
ASPEC & $44.3$        & $24.7$ \hspace{-1.0em} & $(-19.6)$           & $43.5$ \hspace{-1.0em} & $(-0.8)$  & $28.7$        & $18.3$ \hspace{-1.0em} & $(-10.4)$           & $29.3$ \hspace{-1.0em} & $(+0.6)$ \\
JESC  & $14.5$        & $6.6$ \hspace{-1.0em} & $(-7.9)$           & $13.6$ \hspace{-1.0em} & $(-0.9)$    & $17.8$        & $7.0$ \hspace{-1.0em} & $(-10.8)$            & $17.3$ \hspace{-1.0em} & $(-0.5)$ \\
KFTT  & $31.8$        & $17.1$ \hspace{-1.0em} & $(-14.7)$           & $33.0$ \hspace{-1.0em} & $(+1.2)$  & $23.4$        & $13.7$ \hspace{-1.0em} & $(-9.7)$           & $25.4$ \hspace{-1.0em} & $(+2.0)$ \\
IWSLT & $11.1$        & $11.5$ \hspace{-1.0em} & $(+0.4)$           & $14.2$ \hspace{-1.0em} & $(+3.1)$   & $13.7$        & $11.0$ \hspace{-1.0em} & $(-2.7)$           & $17.2$ \hspace{-1.0em} & $(+3.5)$ \\
\bottomrule
\end{tabular}
\caption{BLEU scores of trained NMT models: Differences between in-domain scores are shown in brackets. See Section~\protect\secref{sec:train_with_jparacrawl}\ for in-domain and JParaCrawl columns. For fine-tuning column, see Section~\protect\secref{sec:pretrain_models}.}\label{tab:experimental_results}
\end{table*}

\begin{table*}[t]
\centering
\begin{tabular}{lrrrrrr}
\toprule
\multicolumn{1}{l}{\bf Test excluded} & \multicolumn{3}{c}{\bf English-to-Japanese} & \multicolumn{3}{c}{\bf Japanese-to-English}\\
\cmidrule(lr){2-4}
\cmidrule(lr){5-7}&
\multicolumn{1}{r}{\bf Particular dataset only} & \multicolumn{2}{r}{\bf JParaCrawl} &
\multicolumn{1}{r}{\bf Particular dataset only} & \multicolumn{2}{r}{\bf JParaCrawl} \\ \midrule
ASPEC & $7.9$        & $14.6$ \hspace{-1.0em} & $(+6.7)$   & $5.7$        & $13.8$ \hspace{-1.0em} & $(+8.1)$  \\
JESC  & $5.4$        & $20.1$ \hspace{-1.0em} & $(+14.7)$  & $8.6$        & $16.5$ \hspace{-1.0em} & $(+7.9)$      \\
KFTT  & $4.6$        & $16.8$ \hspace{-1.0em} & $(+12.2)$  & $5.7$        & $14.7$ \hspace{-1.0em} & $(+9.0)$     \\
IWSLT & $5.0$        & $18.5$ \hspace{-1.0em} & $(+13.5)$  & $3.7$        & $14.6$ \hspace{-1.0em} & $(+10.9)$     \\
\bottomrule
\end{tabular}
\caption{BLEU scores of out-of-domain test sets}\label{tab:experimental_results_ood}
\end{table*}

Table~\ref{tab:experimental_results} shows the BLEU scores of the in-domain and JParaCrawl NMT models (see in-domain and JParaCrawl columns).
Since JParaCrawl is not supposed to contain a specific domain, its BLEU scores were lower than those of the model trained with the in-domain corpus.
However, we expect that these in-domain models only focus on a specific domain and do not work well with out-of-domain data.
To test this, we created four out-of-domain test sets by selecting the first 1,000 sentences from each domain (ASPEC, JESC, KFTT, IWSLT) except for one set of in-domain data, resulting in 3,000 sentences for each test set, and measured the BLEU scores.
Table~\ref{tab:experimental_results_ood} shows the BLEU scores of the out-of-domain test set.
As we expected, the models trained with domain specific corpora did not perform well with the out-of-domain test sets; the JParaCrawl model achieved better results.
This means that JParaCrawl contained a broader range of domains.
Thus, a model trained with it might work well as a pre-trained model for fine-tuning with the domain-specific corpora, which we discuss in the following section.

\subsection{Fine-Tuning with Pre-Trained Models}
\label{sec:pretrain_models}
Next, we consider a situation in which we want a domain-specific NMT model with low computational cost.
We fine-tuned the JParaCrawl pre-trained NMT model with a specific domain corpus for domain adaptation.
Such a situation frequently occurs, especially in practical use, since training NMT from the initial state requires huge computational resources.
Below, we address whether we can achieve a comparable performance while reducing the training time.

\subsubsection{Experimental Settings}
The domain-specific corpus we used for fine-tuning is the same as that described in Section~\secref{sec:train_with_jparacrawl_settings}.
Table~\ref{tab:train_set_corpus_stats} shows the corpus statistics.

We started the training from the last saved model trained with JParaCrawl, as described in Section~\secref{sec:train_with_jparacrawl}. Then we further trained the model for 2,000 iterations with the domain-specific corpus.
For the small settings, we changed the number of mini-batches for updates to 16, since the IWSLT corpus is too small to accumulate a large number of mini-batches.
We kept the other settings identical to those described in Section~\secref{sec:train_with_jparacrawl_settings}.
We trained the models on eight NVIDIA RTX 2080 Ti GPUs.
For evaluation, we measured the BLEU scores as well as the training time.

\subsubsection{Experimental Results and Analysis}

\begin{table}[t]
\centering
\begin{tabular}{llrr}
\toprule
\bf Data & \bf Model & \begin{tabular}[b]{r} \bf Without\\\bf pre-training\end{tabular} & \bf Fine-tuning\\ \midrule
ASPEC & big & $13.22$        & $1.60$                    \\
JESC  & big & $14.19$        & $1.69$                    \\
KFTT  & base & $5.65$        & $0.59$                    \\
IWSLT & small & $0.44$        & $0.20$                  \\ \midrule
JParaCrawl & big & $19.37$        & ---                  \\
JParaCrawl & base & $7.36$        & ---                  \\
JParaCrawl & small & $6.83$        & ---                  \\
\bottomrule
\end{tabular}
\caption{Time required (hrs) to train NMT model for English-to-Japanese}
\label{tab:experimental_results_enja_training_time}
\end{table}

Table~\ref{tab:experimental_results} shows the BLEU scores of the fine-tuned models (fine-tuning columns).
Compared to a model trained just with JParaCrawl, a significant performance gain can be seen from the fine-tuning for all the settings.
The ASPEC and JESC experiments show that our fine-tuned models achieved almost comparable performance to the model trained with the in-domain data or even surpassed the ASPEC English-Japanese model.
Our fine-tuned models also surpassed the KFTT and IWSLT settings for both directions.
ASPEC and JESC are already large enough to train solely with the corpus, although smaller corpora such as KFTT and IWSLT have room for improvement by our fine-tuning approach.

Table~\ref{tab:experimental_results_enja_training_time} shows the time required to train a model.
For the fine-tuning experiments, we did not include the time for pre-training the model with JParaCrawl.
Our fine-tuning approach drastically reduced the training time compared to training from the initial state.
These results showed a domain-specific NMT model can be trained in a few hours starting from the pre-trained models with a parallel corpus of the domain.

As shown in this section, the JParaCrawl pre-trained model reduced the training time while maintaining (or boosting) the performance.
This confirms that JParaCrawl can be useful as a pre-training corpus.
We freely provided online the pre-trained models that we used for the experiments\footnote{\url{http://www.kecl.ntt.co.jp/icl/lirg/jparacrawl/}}.

\subsection{Training NMT for a Specific Domain with JParaCrawl}
\label{sec:train_specific_domain}
In the previous section, we discussed the pre-training and fine-tuning approaches to train domain-specific NMT models with reduced computational time.
Next, we focus on the performance on a specific domain, scientific paper excerpts, and ignore the computational time to train the model.
We compare a model trained from the initial state with the existing corpus or with JParaCrawl to achieve the best performance.

\subsubsection{Experimental Settings}
We trained a model for translating scientific domains with ASPEC and with JParaCrawl.
Since JParaCrawl is much larger than ASPEC, we oversampled the latter three times and concatenated it to the former, resulting in 14.76 million sentences in the training data.

As an NMT model, we used Transformer with big settings, as described in Section~\secref{sec:train_with_jparacrawl_settings}.
All hyper-parameters and training/evaluation procedures were identical to those in Section~\secref{sec:train_with_jparacrawl_settings}, except we changed the number of iterations to 25,000 based on the validation perplexity.

\subsubsection{Experimental Results and Analysis}

\begin{table}[t]
\centering
\begin{tabular}{lrr}
\toprule
\textbf{Data}  &  \textbf{En-Ja} & \textbf{Ja-En}\\ \midrule
 ASPEC only           & $44.3$ & $28.7$ \\
 JParaCrawl only           & $24.7$ & $18.3$ \\
 JParaCrawl $\rightarrow$ ASPEC fine-tune & $43.5$ & $29.3$ \\
 JParaCrawl + ASPEC oversample & $44.3$ & $29.8$ \\
 \bottomrule
\end{tabular}
\caption{BLEU scores on ASPEC test set}
\label{tab:aspec_jparacrawl}
\end{table}

Table~\ref{tab:aspec_jparacrawl} shows the experimental results on the ASPEC corpus.
The first and second rows correspond to Section~\secref{sec:train_with_jparacrawl}, and the third row corresponds to Section~\secref{sec:pretrain_models}.
The last row shows the result trained with oversampled ASPEC and JParaCrawl, as described above.

For both directions, the oversampled model outperformed the fine-tuning approach.
In particular, in the Japanese-to-English experiments our model surpassed the model trained only with ASPEC.

These results show that adding JParaCrawl to the existing corpus improved the performance of a specific domain.
To achieve the best translation performance, training should start from the initial state, though the computational time required for this is much greater than that for the fine-tuning approach.

\subsection{Translation Example}
\label{sec:example}

\begin{table*}[t]
\centering
\small
\scalebox{1.0}{
\begin{tabular}{ll}
\toprule
Source & \ja{角柱を過ぎる粘性流体の乱流をラージエディシミュレーションし，}\\
       & \ja{フイルタ幅と数値粘性の影響を調べた。}\\ \midrule
Reference & \uwave{The large eddy simulation} of a turbulent flow of a viscous fluid passing through a \\
& square column was conducted, and the effects of the filter width and numerical \\
& viscosity were examined. \\ \midrule
In-domain only & Turbulent flow of a viscous fluid past a square cylinder was simulated by \uwave{a large} \\
& \uwave{eddy simulation} to examine the effects of filter width and numerical viscosity. \\ \midrule
JParaCrawl only & We simulated \uwave{large-scale} turbulence of viscous fluids past the square column, and \\
& investigated the influence of film width and numerical viscosity. \\ \midrule
JParaCrawl $\rightarrow$ ASPEC fine-tune & Turbulent flow of viscous fluid past a square cylinder was simulated by \uwave{large eddy} \\
& \uwave{simulation}, and the effects of filter width and numerical viscosity were examined. \\ \midrule
JParaCrawl + ASPEC (without pre-training) & Turbulent flow of viscous fluid past a square cylinder is simulated by \uwave{large eddy} \\
& \uwave{simulation}, and the effects of filter width and numerical viscosity are investigated. \\
\bottomrule
\end{tabular}
}
\caption{Example translations of trained models: The translation domain is scientific paper excerpts.}
\label{tab:example_translation}
\end{table*}

To identify the strengths and weaknesses of the models, we compared the translations of the models trained for the previous experiments.
Table~\ref{tab:example_translation} shows example translations from a domain of scientific paper excerpts.

Our results show that the JParaCrawl model sometimes made mistakes when translating domain-specific words.
This example includes ``The large eddy simulation,'' which is an uncommon, domain-specific expression.
The model trained with the in-domain corpus correctly translated it; however, the model trained with JParaCrawl only mistranslated it as ``large-scale.''
The output of the JParaCrawl model is somewhat understandable, but it does not work as accurately as the one trained with an in-domain corpus, in particular when the input is very domain-specific.

However, after fine-tuning with the in-domain corpus, the model can translate these domain-specific words correctly.
This is almost the same as the model trained with JParaCrawl + ASPEC (without pre-training), showing that the pre-training and fine-tuning approach could improve the translation accuracy while only requiring small computational time.

\section{Conclusion}
\label{sec:conclusion}
We introduced JParaCrawl, a large web-based English-Japanese parallel corpus.
It was made by crawling the web and finding English-Japanese bitexts.
After filtering out possibly noisy sentences, we retained around 8.7 million parallel sentences, which we publicly released as the corpus.

Our experiments showed how JParaCrawl contains a broader range of domains and can be used for general purposes.
We also drastically reduced the training time by fine-tuning the JParaCrawl pre-trained NMT models and in doing so maintained or even boosted the performance.
Finally, we showed that JParaCrawl also improved the performance of a specific domain when training a model with an existing corpus from an initial state.

In future work, we will crawl more websites and make the dataset larger.
We also plan to improve the bitext aligner and cleaner, especially for Japanese.
We only focused on English-Japanese in the initial release, but we hope to eventually add more language pairs to/from Japanese.

\section{Acknowledgements}
We gratefully acknowledge the ParaCrawl project for releasing the software and useful discussions.
We thank the three anonymous reviewers for their useful comments.
We would also like to thank Hisashi Itoh and Takumi Asai for their technical support.

\section{Bibliographical References}
\label{sec:ref}

\bibliographystyle{lrec}
\bibliography{myplain,main}

\end{document}